\documentclass[letterpaper]{article}
\usepackage{iclr2027_conference,times}
\usepackage[hyphens]{url}
\usepackage{graphicx}
\usepackage{xcolor}
\usepackage{natbib}
\usepackage{caption}
\usepackage{amsmath}
\usepackage{amssymb}
\usepackage{amsthm}
\newtheorem{proposition}{Proposition}
\usepackage{booktabs}
\usepackage{array}
\usepackage{colortbl}
\usepackage{multirow}
\usepackage{placeins}
\usepackage{float}
\usepackage{algorithm}
\usepackage{algpseudocode}
\usepackage{titletoc}

\usepackage{hyperref}
\definecolor{CitationBlue}{HTML}{245A81}
\hypersetup{
  colorlinks=true,
  citecolor=CitationBlue,
  urlcolor=CitationBlue,
  linkcolor=black,
  linktoc=all,
  pdftitle={Large Language Models without QK-Normalization},
  pdfauthor={Yongding Tao, Yitong Qiao, Tianqi Shao, Tian Wang, Zhixuan Chu, Xiaoming Wu, Yihong Dong, Ge Li}
}
\iclrfinalcopy

\newcommand{\method}{QK-MAP}
\newcommand{\qknorm}{QK-Norm}
\newcommand{\qkclip}{QK-Clip}
\newcommand{\mapdim}{QK-MAP-D}
\newcommand{\mapparam}{QK-MAP-P}
\newcommand{\mapres}{QK-MAP-R}

\definecolor{OLMOHeader}{HTML}{E9ECEF}
\definecolor{OLMOMethod}{HTML}{EEF4FA}
\definecolor{OLMOSubtle}{HTML}{F7F7F7}
\definecolor{OLMOSpine}{HTML}{707070}

\newcommand{\thead}[1]{\textbf{#1}}
\newcommand{\tableheadrow}{}
\newcommand{\methodrow}{\rowcolor{OLMOMethod}}
\newcommand{\subtlerow}{\rowcolor{OLMOSubtle}}
\newcommand{\methodcell}[1]{\cellcolor{OLMOMethod}#1}
\newcommand{\datatablestyle}{  \small
  \renewcommand{\arraystretch}{1.12}  \setlength{\tabcolsep}{4pt}  \arrayrulecolor{OLMOSpine}}
\newcommand{\compactdatatablestyle}{  \small
  \renewcommand{\arraystretch}{1.08}  \setlength{\tabcolsep}{2.0pt}  \arrayrulecolor{OLMOSpine}}
\newsavebox{\iclrtablebox}
\newenvironment{fittable}
  {\begin{lrbox}{\iclrtablebox}}
  {\end{lrbox}   \ifdim\wd\iclrtablebox>\linewidth
     \resizebox{\linewidth}{!}{\usebox{\iclrtablebox}}   \else
     \usebox{\iclrtablebox}   \fi}
\newenvironment{appendixwidetable}
  {\begin{table}[!tbp]\centering}
  {\end{table}}
\newenvironment{appendixwidefigure}
  {\begin{figure}[!tbp]\centering}
  {\end{figure}}

\title{Large Language Models\\without QK-Normalization}
\author{\normalfont\textbf{Yongding Tao}$^{1,2}$\quad \textbf{Yitong Qiao}$^{3}$\quad
\textbf{Tianqi Shao}$^{1,2}$\quad \textbf{Tian Wang}$^{1,2}$\\
\textbf{Zhixuan Chu}$^{3}$\quad \textbf{Xiaoming Wu}$^{4}$\quad
\textbf{Yihong Dong}$^{5}$\quad \textbf{Ge Li}$^{1,2}$\\
$^1$ School of Computer Science, Peking University\\
$^2$ Key Laboratory of High Confidence Software Technologies, Ministry of Education\\
$^3$ Zhejiang University \quad $^4$ Qilu University of Technology\\
$^5$ Shanghai Jiaotong University\\
\texttt{ydtao25@stu.pku.edu.cn} \quad \texttt{lige@pku.edu.cn}
}

\begin{document}
\maketitle

\begin{abstract}
Query-Key Normalization (QK-Norm) is widely adopted in modern large
language models to control the growth of
attention logits and improve training stability. However, query and key
norms are not merely sources of instability. The query norm controls the
sharpness of the attention distribution, while the key norm determines
how strongly a token attracts different queries. These norms therefore
carry token-specific information about attention selectivity and token
importance. By normalizing both, QK-Norm suppresses this information.
This raises a natural question: \textit{can we control query-key
representations without discarding their sensitivity to magnitude?}
To address this question, we propose \textbf{QK-MAP}, a
\textbf{M}agnitude-\textbf{A}ware \textbf{P}rojection for constructing
queries and keys. Rather than removing magnitude through normalization,
QK-MAP transforms the input representation using multiple learnable
frequencies, so that changes in magnitude are encoded as bounded phase
patterns. This allows magnitude to be represented through phase and
direction rather than vector length alone. Its periodic features have a
fixed norm while remaining sensitive to magnitude changes, separating
magnitude encoding from norm growth. Multiple frequencies further help
distinguish magnitudes and reduce periodic ambiguity.
Experiments across model scales from 190M to 7B parameters show that
QK-MAP achieves lower training and validation loss than both Standard
attention and QK-Norm. Further analyses show that QK-MAP preserves
sensitivity to input scale, assigns less attention to the first token, and
produces stronger attention contrast for relevant evidence in synthetic
long-context tests. These results demonstrate that query-key scale can be
effectively controlled without sacrificing token-specific magnitude
information.
Our code is available at \url{https://github.com/yongding-tao/QK-MAP}.
\end{abstract}

\section{Introduction}

Query-Key Normalization (QK-Norm) has become a common component in
modern large language models, including OLMo~3 and Qwen~3
\citep{teamolmo2025olmo3,yang2025qwen3}.
It addresses a common training problem: query and key norms can grow
during training and produce increasingly large attention logits.
Large logits can saturate the softmax, reduce attention entropy, and
make optimization unstable
\citep{dehghani2023scaling,zhai2023entropy,wortsman2023proxies}.
QK-Norm controls this growth by normalizing projected queries and keys
before computing their dot products
\citep{henry2020qknorm,rybakov2024stability}.
This prevents attention logits from growing directly with query and key
norms and provides a practical way to improve training stability.

\begin{figure}[!tbp]
\centering
\includegraphics[width=\linewidth]{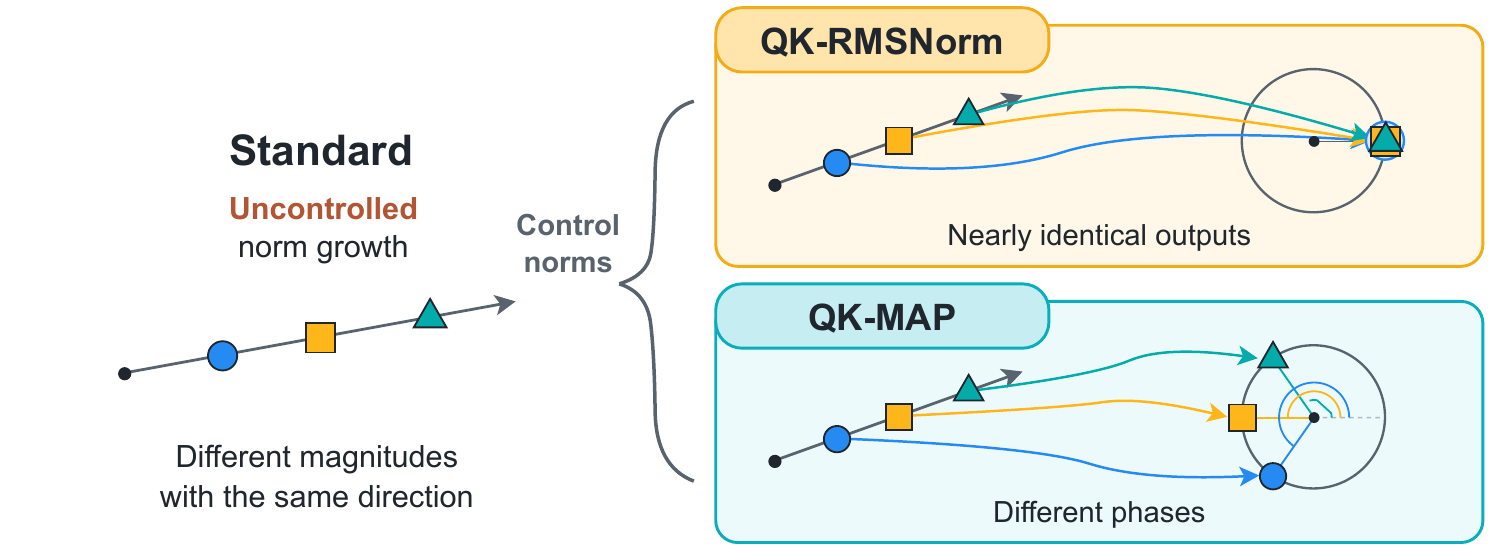}
\caption{
Schematic comparison for inputs with the same direction and different
magnitudes. Standard attention preserves magnitude differences but allows
query-key norms to grow. QK-RMSNorm controls norms but maps these inputs
to nearly the same representation. QK-MAP's periodic features change with
input magnitude while keeping their norm fixed.
}
\label{fig:motivation}
\end{figure}

However, query and key norms do more than control the size of attention
logits.
They are also part of the attention function.
For a query $q_i$ and key $k_j$,
\begin{equation}
q_i^\top k_j
=
\|q_i\|_2\|k_j\|_2\cos\theta_{ij}.
\label{eq:norm-direction}
\end{equation}
The angle $\theta_{ij}$ describes how well the query and key directions
match, while the two norms control the strength of this match.
The query norm scales an entire row of attention logits.
It therefore acts as a query-dependent temperature: a larger query norm
generally produces a sharper attention distribution.
The key norm scales a column of logits and changes one token's scores
across different queries.
The two norms thus play different roles and carry token-specific
information within attention.

Controlling these norms therefore involves a choice.
Standard attention preserves this token-specific magnitude information,
but allows query and key norms to grow.
QK-Norm controls this growth, but makes the resulting representations
much less sensitive to magnitude.
This motivates the main question of this work:
\emph{Can we control query-key representations without losing their
sensitivity to magnitude?}

To address this question, we propose \textbf{QK-MAP}, a
\textbf{M}agnitude-\textbf{A}ware \textbf{P}rojection for constructing
queries and keys.
QK-MAP first maps an input representation into periodic features and
then uses these features to construct queries and keys.
When the input magnitude changes, the phases and coordinates of these
features change while their overall norm remains fixed.
Magnitude can therefore be represented by the pattern of the
representation rather than only by its vector length.
In this way, QK-MAP keeps query-key representations sensitive to
magnitude without requiring their feature norms to grow with the input.

Figure~\ref{fig:motivation} illustrates the main idea.
Consider several inputs that have the same direction but different
magnitudes.
After QK-Norm, they produce nearly the same representation.
With QK-MAP, changing the input magnitude changes the phase and
coordinate pattern of the representation, even though its feature norm
does not change.
QK-MAP therefore moves magnitude information from vector length into
the representation pattern.

We evaluate QK-MAP in language models ranging from 190M to 7B
parameters under closely matched pretraining settings.
Across model scales, QK-MAP achieves lower training and validation loss
than both Standard attention and QK-Norm.
At 7B parameters, it also obtains the best average accuracy on the main
eight-task evaluation suite.
Our analyses further show that QK-MAP retains sensitivity to input
scale, assigns less attention to the first token, and gives relevant
evidence stronger attention contrast in synthetic long-context tests.
These results show that controlling query-key scale does not require
removing token-specific magnitude information.

Our contributions are:
\begin{itemize}
    \item We explain the different roles of query and key norms in
    attention and identify a tradeoff in QK-Norm: it controls logit
    growth but also removes much of the token-specific magnitude
    information carried by queries and keys.

    \item We introduce QK-MAP, which represents changes in input
    magnitude through fixed-norm phase and coordinate patterns.
    This keeps query-key representations sensitive to magnitude
    without relying only on growth in vector length.

    \item We evaluate QK-MAP from 190M to 7B parameters.
    The results show lower training and validation loss, favorable
    downstream performance, and clear changes in scale sensitivity,
    positional attention allocation, and relevant-evidence contrast.
\end{itemize}

\section{Magnitude Information in Attention}
\label{sec:background}

\subsection{Magnitude as an Attention Signal}

For one attention head with width $d_h$, let
\begin{equation}
q_i=x_iW_Q,\qquad
k_j=x_jW_K,\qquad
S_{ij}=\frac{q_i^\top k_j}{\sqrt{d_h}}.
\end{equation}
The attention distribution for query $i$ is
$a_i=\operatorname{softmax}(S_{i,:})$.
Writing
$q_i=r_i^q u_i^q$ and $k_j=r_j^k u_j^k$,
where $u_i^q$ and $u_j^k$ have unit norm, gives
\begin{equation}
S_{ij}
=
\frac{r_i^q r_j^k}{\sqrt{d_h}}
(u_i^q)^\top u_j^k.
\label{eq:decomposed-score}
\end{equation}

The directional term measures how well a query and key match, while
their magnitudes control how strongly this match is expressed.
For fixed keys, $r_i^q$ scales the entire $i$-th row of the logit
matrix.
It therefore acts as a query-dependent inverse temperature:
increasing $r_i^q$ makes the attention distribution sharper unless all
logits are equal.
For fixed queries, $r_j^k$ scales the $j$-th column and changes token
$j$'s scores across different queries.
Query and key magnitudes therefore carry different token-specific
signals, although either can also produce overly large logits.

These two effects are complementary.
Query magnitude controls how selectively a token reads from context,
while key magnitude controls how strongly that token is considered by
other queries.
QK-Norm suppresses both forms of token-specific control.

\subsection{Norm Control versus Magnitude Sensitivity}

A direct way to control query and key norms is $\ell_2$ normalization:
\begin{equation}
\mathcal{L}(z)=\frac{z}{\|z\|_2}.
\label{eq:l2norm}
\end{equation}
It fixes the output norm to one, preventing the input magnitude from
directly increasing the attention logits.
However, for any fixed nonzero direction $u$ and $r>0$,
\begin{equation}
\mathcal{L}(ru)=\mathcal{L}(u).
\end{equation}
$\ell_2$ normalization therefore controls norm growth by making all positive
rescalings along the same direction indistinguishable.

Modern language models commonly use RMSNorm instead of exact $\ell_2$
normalization:
\begin{equation}
\mathcal{N}_{\epsilon}(z)
=
\frac{z}{\sqrt{\|z\|_2^2/d_h+\epsilon}}\odot g,
\label{eq:rmsnorm}
\end{equation}
where $g$ is a learned gain shared across tokens
\citep{zhang2019rmsnorm}.
When $\epsilon=0$ and the gain is ignored,
$\mathcal{N}_{0}(z)$ differs from $\ell_2$ normalization only by the constant
factor $\sqrt{d_h}$.
It therefore has the same invariance to input magnitude.

\begin{proposition}[Magnitude information after QK-RMSNorm]
\label{prop:qknorm-scale}
Let $z=ru$, where $u\neq 0$ is fixed and $r>0$.
For $\epsilon=0$,
\begin{equation}
\mathcal{N}_{0}(ru)=\mathcal{N}_{0}(u).
\end{equation}
For $\epsilon>0$,
\begin{equation}
\left\|
\frac{\partial\mathcal{N}_{\epsilon}(ru)}{\partial r}
\right\|_2
=
\frac{\epsilon\|u\|_2}
{(r^2c+\epsilon)^{3/2}},
\qquad
c=\frac{\|u\|_2^2}{d_h},
\label{eq:qknorm-scale-derivative}
\end{equation}
where the fixed gain $g$ is omitted for clarity.
The dependence on magnitude therefore decreases rapidly as $r$ grows.
\end{proposition}

QK-L2Norm fixes the norm exactly, while QK-RMSNorm allows learned
dimension-wise scaling through $g$ and uses a finite $\epsilon$ for
numerical stability.
Despite this difference, both prevent token magnitude from directly
controlling the query or key norm.
The gain $g$ is shared across tokens, so it cannot restore a different
magnitude signal for each query or key.
Thus, the same mechanism that controls token-wise norm growth also
removes most of the information carried by changes in magnitude.

\begin{figure}[htbp]
\centering
\includegraphics[width=\textwidth]
{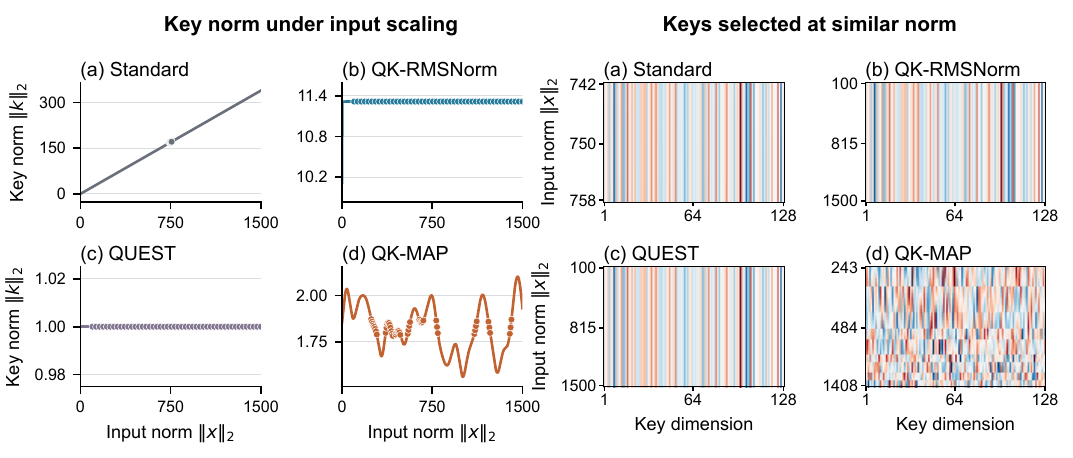}
\caption{
Key-side response to input scaling.
\emph{Left:} Key norms as input norm varies; dots mark samples within a
narrow, method-specific key-norm range.
\emph{Right:} Key coordinates of the selected samples, ordered by input
norm. Each heatmap uses its own color scale.
}
\label{fig:scale-probe}
\end{figure}

QUEST \citep{govindarajan2026quest} takes an asymmetric approach: it
normalizes keys to unit length but leaves queries unnormalized.
It therefore retains query-dependent attention sharpness while fixing
the key-norm factor in Equation~\ref{eq:decomposed-score}.
We include QUEST in the key-side comparison in Figure~\ref{fig:scale-probe}.

\section{QK-MAP}

\label{sec:qkmap}

QK-MAP replaces QK normalization with a learned periodic
representation.
Its goal is to keep the query-key path responsive to input magnitude
without requiring that information to be carried only by vector length.

Let $x\in\mathbb{R}^{D}$ be the representation entering the attention
block.
QK-MAP learns
$W_p\in\mathbb{R}^{D\times m}$ and
$b_p\in\mathbb{R}^{m}$, and computes
\begin{equation}
\Phi(x)=
\begin{bmatrix}
\cos(xW_p+b_p)\\
\sin(xW_p+b_p)
\end{bmatrix}
\in\mathbb{R}^{2m}.
\label{eq:feature-map}
\end{equation}
The sine and cosine are applied elementwise.
In our main configuration, $m=D/2$, so the output dimension of
$\Phi(x)$ is also $D$.

\begin{figure}[t]
\centering
\includegraphics[width=0.9\textwidth]
{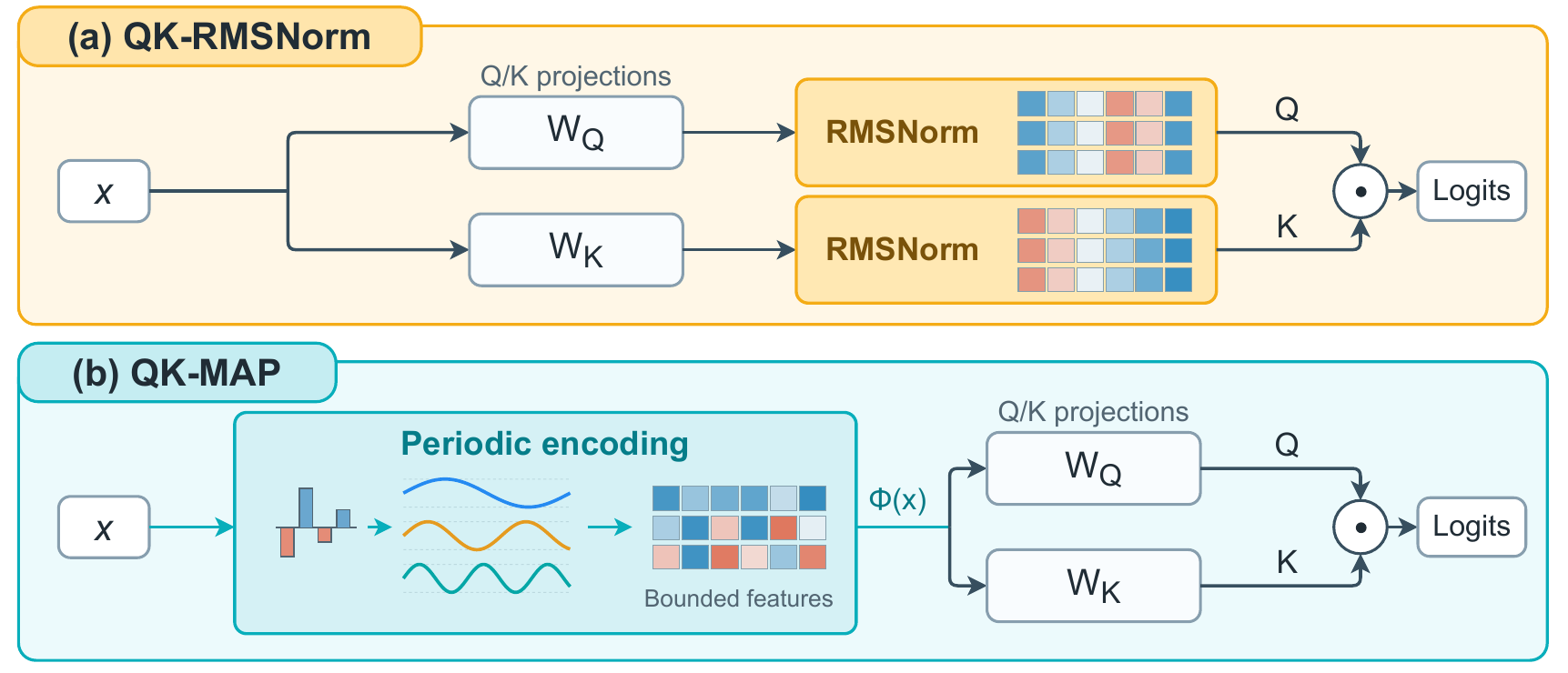}
\caption{
Comparison of QK-RMSNorm and QK-MAP.
QK-RMSNorm normalizes projected queries and keys before attention.
QK-MAP first forms a shared periodic representation and then applies
separate query and key projections.
}
\label{fig:framework}
\end{figure}

The same periodic representation is used to form queries and keys,
with separate projections for each path:
\begin{equation}
q=\Phi(x)W_Q,\qquad
k=\Phi(x)W_K,\qquad
v=xW_V.
\label{eq:qkmap-attention}
\end{equation}
The value path is unchanged. After projection, queries and keys are
reshaped into heads, and RoPE is applied as usual. The frequencies in
$W_p$ are learned jointly with the language model.
Figure~\ref{fig:framework} compares where the two methods control
query-key scale.

The design is inspired by the representation idea behind RoPE
\citep{su2024roformer}.
RoPE represents position through rotations at multiple base
frequencies: each coordinate remains bounded, while different
positions produce different combinations of phases over the context
range.
QK-MAP applies a similar idea to magnitude.
Instead of using fixed frequencies to encode a position index, it uses
learned frequencies to turn changes in the input representation into
changes across multiple phase coordinates.
In this way, a potentially unbounded magnitude can be represented by a
bounded pattern.
This use differs from random and learned Fourier features for kernel
approximation or coordinate encoding
\citep{rahimi2007random,tancik2020fourier,li2021learnablefourier}.
Here the periodic map is learned inside each attention block and is
applied to the token representation rather than to a position index.

The key property is stated below.

\begin{proposition}[Magnitude encoding without norm growth]
\label{prop:qkmap-scale}
For the feature map in Equation~\ref{eq:feature-map},
\begin{equation}
\|\Phi(x)\|_2^2=m
\qquad
\text{for every }x.
\label{eq:fixed-feature-norm}
\end{equation}
For a fixed direction $u$ and $x=ru$,
\begin{equation}
\left\|
\frac{\partial\Phi(ru)}{\partial r}
\right\|_2^2
=
\|uW_p\|_2^2.
\label{eq:qkmap-scale-response}
\end{equation}
Therefore, the feature norm remains fixed, while the feature values
continue to change with input magnitude whenever $uW_p\neq 0$.
\end{proposition}

This result gives the main difference from QK-Norm.
QK-Norm makes representations with the same direction nearly
independent of their magnitude.
QK-MAP instead moves magnitude information from vector length into the
coordinates of the representation.

The use of multiple frequencies also matters.
A single phase coordinate repeats as the input magnitude changes.
With multiple frequencies, the complete feature vector repeats only
when all phase coordinates repeat at the same time.
For a fixed direction $u$, two magnitudes $r_1\neq r_2$ produce the
same periodic features only if
\begin{equation}
(r_1-r_2)u^\top w_j\in 2\pi\mathbb{Z}
\qquad
\text{for every }j.
\label{eq:same-features}
\end{equation}
Each phase coordinate is periodic, but multiple learned frequencies provide a
richer collection of phase patterns through which the model can
represent magnitude.

Figure~\ref{fig:scale-probe} illustrates these properties on the key
side so that QUEST, which normalizes only keys, can be compared directly.
For a fixed direction $u$, Standard key norms grow with the input
magnitude $r$.
QK-RMSNorm and QUEST keep the key norm nearly constant, whereas QK-MAP
keeps it bounded while still responding to $r$.
The right block compares inputs whose key norms are similar within each
method.
The normalized baselines give nearly identical key coordinates across
widely separated input magnitudes.
QK-MAP instead gives different coordinate patterns even when its key
norms are similar.
Thus, controlling key norm need not remove all information about input
magnitude.

\section{Experiments}
\label{sec:experiments}

\subsection{Setup}

We compare Standard attention, \qknorm{}, and \method{} in
decoder-only language models from 190M to 7B parameters.
Within each scale, all variants use the same backbone, pretraining
data, token budget, global batch size, and initialization and data seeds.
The pretraining data follow the Dolma-based OLMo mixture
\citep{groeneveld2024olmo,soldaini2024dolma}.
The cross-scale comparison uses a learned per-dimension residual variant
of \method{}, with a slightly reduced feed-forward width to match
parameter counts.
Architecture and residual ablations are in Appendix~\ref{app:ablations}.

For downstream evaluation, we follow the OLMES protocol
\citep{gu2025olmes} and report Avg-8, the unweighted mean accuracy over
ARC-Easy \citep{clark2018arc}, PIQA \citep{bisk2020piqa},
SciQ \citep{welbl2017sciq}, ARC-Challenge \citep{clark2018arc},
COPA \citep{roemmele2011copa}, HellaSwag \citep{zellers2019hellaswag},
OpenBookQA \citep{mihaylov2018openbookqa}, and
WinoGrande \citep{sakaguchi2021winogrande}.
For language modeling, we report final-window training cross-entropy
and validation cross-entropy on C4 \citep{raffel2020t5},
Pile \citep{gao2020pile}, M2D2-S2ORC \citep{reid2022m2d2},
and six Dolma domains \citep{soldaini2024dolma}.
We then examine magnitude use, attention allocation, and performance
under text perturbations.
Full configurations and evaluation protocols are in
Appendices~\ref{app:configuration} and~\ref{app:evaluation}.

\subsection{Main Results}

\begin{table}[H]
    \centering
    \small
    \renewcommand{\arraystretch}{1.02}
    \setlength{\tabcolsep}{2.0pt}
    \arrayrulecolor{OLMOSpine}
\begin{fittable}
    \begin{tabular}{ll*{10}{c}@{\hspace{7pt}}>{\centering\arraybackslash}p{3.4em}}
    \toprule
    \tableheadrow
    \multirow{2}{*}[-2pt]{\thead{Scale}} & \multirow{2}{*}[-2pt]{\thead{Variant}} &
    \multicolumn{2}{c}{\thead{CE$\downarrow$}} &
    \multicolumn{9}{c}{\thead{Downstream evaluation (\%$\uparrow$)}}\\
    \cmidrule(lr){3-4}\cmidrule(lr){5-13}
    \tableheadrow
     &  & \thead{Train} & \thead{Val.} & \thead{ARC-E} & \thead{PIQA} & \thead{SciQ} & \thead{ARC-C} & \thead{COPA} & \thead{HS} & \thead{OBQA} & \thead{WG} & \thead{Avg-8}\\
    \midrule

    \multirow{3}{*}{190M}
     & Standard & 3.86 & 4.19 & 27.72 & 58.10 & 30.30 & 23.75 & 56.00 & 27.40 & 27.80 & 50.90 & 37.75\\
     & \qknorm & 3.78 & 4.10 & \textbf{35.79} & \textbf{60.70} & \textbf{34.10} & \textbf{25.08} & 58.00 & 27.60 & \textbf{30.00} & 48.00 & \textbf{39.91}\\
    \methodrow
     & \method & \textbf{3.67} & \textbf{3.98} & 30.74 & 60.62 & 32.93 & 23.22 & \textbf{61.00} & \textbf{28.20} & 26.33 & \textbf{51.03} & 39.26\\

    \midrule
    \multirow{3}{*}{370M}
     & Standard & 3.19 & 3.54 & 39.12 & 58.90 & 67.00 & \textbf{31.10} & 60.00 & 25.90 & 28.80 & \textbf{52.00} & 45.35\\
     & \qknorm & 3.17 & 3.52 & \textbf{40.70} & \textbf{59.40} & \textbf{70.00} & 23.75 & \textbf{64.00} & 27.00 & 33.60 & 48.80 & 45.91\\
    \methodrow
     & \method & \textbf{3.10} & \textbf{3.45} & \textbf{40.70} & 58.00 & 69.70 & 25.42 & 61.00 & \textbf{27.30} & \textbf{35.60} & 49.80 & \textbf{45.94}\\

    \midrule
    \multirow{3}{*}{760M}
     & Standard & 2.79 & 3.16 & \textbf{47.72} & 62.00 & 79.80 & \textbf{31.44} & 64.00 & 32.30 & 34.40 & \textbf{50.30} & \textbf{50.24}\\
     & \qknorm & 2.79 & 3.16 & 47.19 & 61.70 & 77.50 & 29.77 & \textbf{65.00} & 32.30 & 33.60 & 49.80 & 49.61\\
    \methodrow
     & \method & \textbf{2.76} & \textbf{3.13} & 45.09 & \textbf{62.50} & \textbf{79.90} & 28.43 & 64.00 & \textbf{33.20} & \textbf{34.80} & 49.90 & 49.73\\

    \midrule
    \multirow{3}{*}{1B}
     & Standard & 2.69 & 3.09 & 48.42 & \textbf{63.90} & \textbf{82.00} & 27.76 & 66.00 & 35.00 & 36.00 & \textbf{51.70} & 51.35\\
     & \qknorm & 2.70 & 3.09 & \textbf{51.05} & 63.40 & 79.40 & 31.77 & 66.00 & \textbf{35.70} & \textbf{37.40} & 51.50 & 52.03\\
    \methodrow
     & \method & \textbf{2.67} & \textbf{3.07} & 49.30 & \textbf{63.90} & \textbf{82.00} & \textbf{34.78} & \textbf{67.00} & 34.20 & 34.40 & 51.30 & \textbf{52.11}\\

    \midrule
    \multirow{3}{*}{7B}
     & Standard & 2.25 & 2.89 & 49.65 & \textbf{68.00} & \textbf{84.80} & 37.12 & 69.00 & 46.60 & 39.00 & \textbf{52.70} & 55.86\\
     & \qknorm & 2.24 & 2.90 & \textbf{54.21} & 67.90 & 84.20 & 32.44 & \textbf{72.00} & 45.30 & 37.80 & 50.00 & 55.48\\
    \methodrow
     & \method & \textbf{2.23} & \textbf{2.86} & 52.81 & \textbf{68.00} & 83.00 & \textbf{39.13} & 70.00 & \textbf{47.50} & \textbf{41.20} & 52.40 & \textbf{56.75}\\

    \bottomrule
    \end{tabular}
\end{fittable}
    \caption{
    Language-modeling cross-entropy and downstream accuracy from
    190M to 7B.
    }
    \label{tab:main-results}
\end{table}

\paragraph{Language modeling and downstream performance.}
Table~\ref{tab:main-results} shows that \method{} achieves the lowest
training and validation CE at every evaluated scale.
The validation improvement holds across C4, Pile, M2D2-S2ORC, and
the Dolma domains.
\method{} also improves average downstream accuracy at most model scales,
extending its benefits beyond language modeling loss.

\begin{figure}[!t]
\centering
\includegraphics[width=0.96\linewidth]{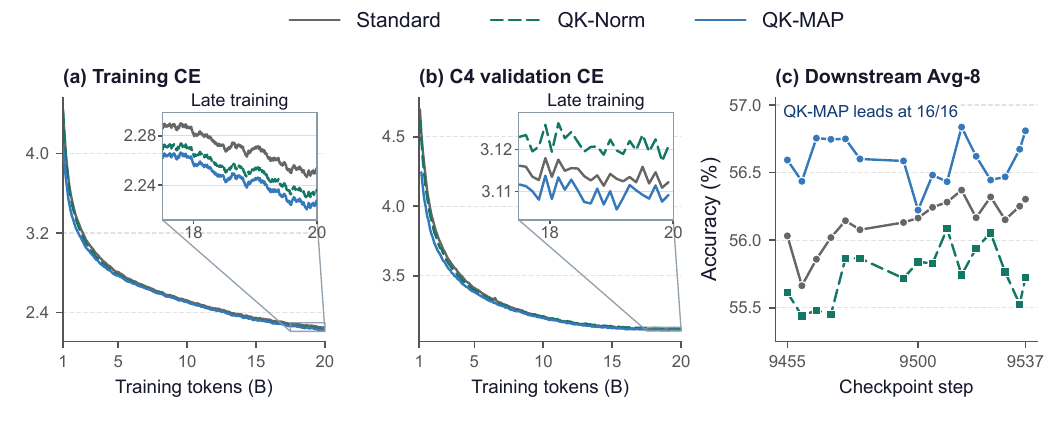}
\caption{Training dynamics of the 7B models.
(a) Training CE with 100-update smoothing.
(b) C4 validation CE, evaluated every 50 updates.
Insets and connecting lines show the final stage of training.
(c) Avg-8 at 16 common checkpoints near the end of training.}
\label{fig:7b-dynamics}
\end{figure}

\paragraph{Training dynamics.}
Figure~\ref{fig:7b-dynamics} shows the training and C4 validation
trajectories at 7B.
The late-training views make the persistent CE advantage of
\method{} visible beyond the final checkpoint.
Panel (c) shows that the downstream result is also consistent across nearby checkpoints:
\method{} ranks first in Avg-8 at all 16 common checkpoints
evaluated near the end of training
(Appendix~\ref{app:checkpoint-robustness}).

\subsection{Do the Models Use Magnitude Information?}
\label{sec:magnitude-intervention}

\begin{figure}[!htbp]
\centering
\includegraphics[width=0.96\linewidth]{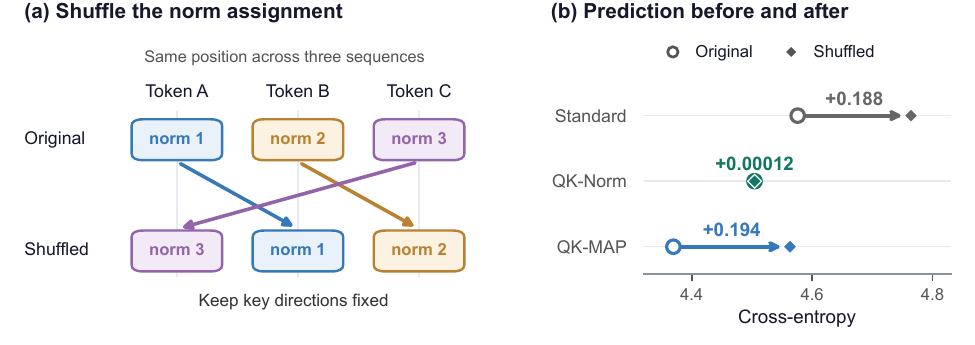}
\caption{Key-norm intervention at 190M.
(a) Reassigning norms among tokens; norm values are illustrative.
(b) CE before and after shuffling, with the increase labeled.}
\label{fig:norm-shuffle}
\end{figure}

If key magnitudes aid prediction, assigning them to different tokens
should increase loss.
We shuffle key norms across sequences at the same position before
attention (Figure~\ref{fig:norm-shuffle}(a)), keeping directions and
the set of norms fixed.
We compare CE before and after shuffling on 128 C4 sequences,
using 190M checkpoints and the direct, same-dimensional \method{} variant.
Appendix~\ref{app:norm-shuffle} gives settings and confidence intervals.

Shuffling key norms raises CE by 0.194 for \method{} and 0.188
for Standard, but only 0.00012 for \qknorm{}
(Figure~\ref{fig:norm-shuffle}(b)).
The loss increase shows that \method{} uses token-specific key
magnitudes for prediction, as Standard attention does.

\subsection{Where Does Attention Go?}
\label{sec:attention-analysis}

We next examine how \method{} allocates attention in the 7B models.
Two complementary measurements track attention to the first key
and attention to inserted evidence.

\begin{figure}[!t]
\centering
\includegraphics[width=0.96\linewidth]{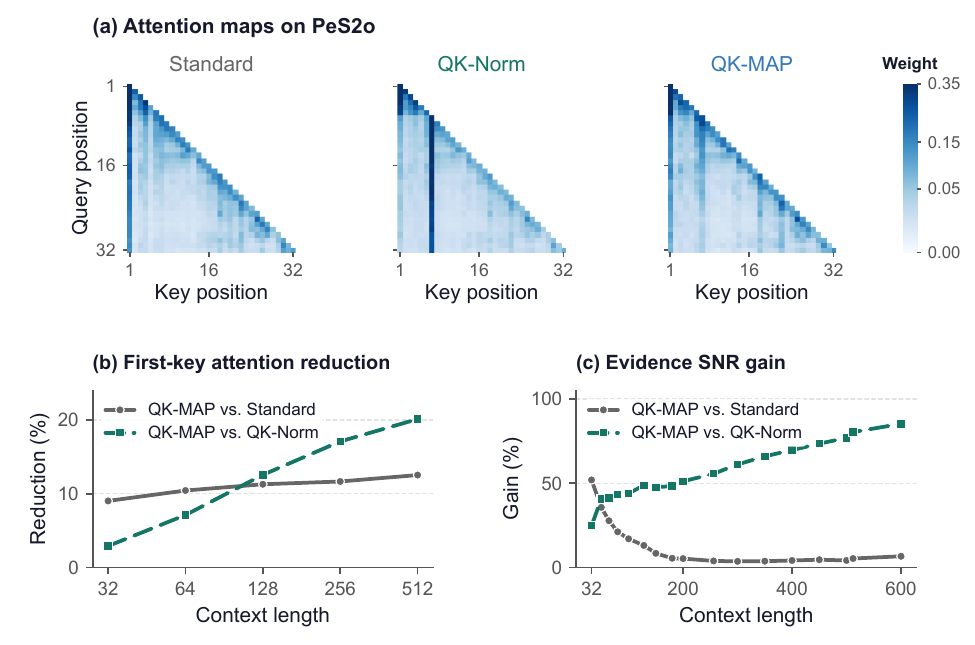}
\caption{Attention allocation in the 7B models.
(a) PeS2o attention maps over the first 32 positions, with a shared
square-root color scale.
(b) Relative reduction in mean first-key attention over eleven domains.
(c) Relative gain in average-layer evidence SNR.
Panels (b)--(c) show \method{} relative to each baseline.}
\label{fig:attention-redistribution}
\end{figure}

\paragraph{Less concentration on the first key.}
Initial tokens often receive large attention weights, a behavior
associated with attention sinks
\citep{xiao2023streamingllm,gu2025attentionsink}.
We measure the percentage of attention assigned to the first key,
averaged over layers, heads, and query positions.
Figure~\ref{fig:attention-redistribution}(a) gives a direct view:
the strong vertical attention bands in the PeS2o baseline maps are
weaker under \method{}.
Panel (b) summarizes the relative reduction in mean first-key
attention over eleven domains at context lengths from 32 to 512.
\method{} has the lowest first-key attention at every length and is
below both baselines in all 55 domain--length comparisons.
Its mean is 9.1\%--12.7\% lower than Standard.
Domain-level results and attention maps are in
Appendix~\ref{app:first-token-allocation}.

\paragraph{Stronger attention to relevant evidence.}
To examine content allocation, we insert evidence into distractor text
and measure evidence-to-distractor attention SNR for the final
question-token query.
We average the statistic across all layers.
Such synthetic prompts offer a controlled setting for studying
attention to relevant content
\citep{liu2024lost,hsieh2024ruler,bai2023longbench,modarressi2025nolima}.
Figure~\ref{fig:attention-redistribution}(c) shows that \method{}
has the highest average-layer SNR at every evaluated context length.
Its relative gain over \qknorm{} grows from 24.9\% at 32 tokens
to 85.3\% at 600 tokens.
The pattern also holds when the prompt contains more evidence or
different task types, as shown in Table~\ref{tab:niah-average-layer}.

\begin{table}[!htbp]
\centering
\small
\renewcommand{\arraystretch}{1.10}
\setlength{\tabcolsep}{17pt}
\begin{tabular}{lccc}
\toprule
\tableheadrow
\thead{Prompt set} & \thead{Standard} & \thead{\qknorm} & \thead{\method}\\
\midrule
1 needle   & 0.272 & 0.238 & \methodcell{\textbf{0.338}}\\
2 needles  & 0.352 & 0.318 & \methodcell{\textbf{0.426}}\\
3 needles  & 0.418 & 0.415 & \methodcell{\textbf{0.506}}\\
Multi-task & 0.342 & 0.290 & \methodcell{\textbf{0.386}}\\
\bottomrule
\end{tabular}
\caption{Average-layer evidence attention SNR at 7B.
Higher values indicate stronger contrast between evidence and
distractors.}
\label{tab:niah-average-layer}
\end{table}

Together, these results show two consistent changes:
\method{} assigns less attention to the first key and gives
relevant evidence stronger attention contrast.
The evidence result extends from single-needle prompts to multiple
needles and mixed tasks, rather than depending on one prompt format.
Appendix~\ref{app:evidence-contrast} gives the prompt settings and
additional layer-pooling results.

\subsection{Performance under Text Perturbations}
\label{sec:text-perturbation}

We also test whether the language-modeling advantage persists when
the input text is perturbed.
Following the word-replacement test used by QUEST
\citep{govindarajan2026quest}, we randomly replace words with
\texttt{AAA} in a fixed WikiText-103 validation sample
\citep{merity2016wikitext}.
We evaluate Standard, \qknorm{}, QUEST, and the direct \method{}
variant at 190M using identical text and replacements.
The replacement rates are 1.5\%, 2.5\%, and 5\%;
Appendix~\ref{app:text-perturbation} gives the evaluation protocol.

\begin{table}[!htbp]
\noindent
\begin{minipage}[t]{0.51\linewidth}
\vspace{0pt}
\centering
\small
\renewcommand{\arraystretch}{1.16}
\setlength{\tabcolsep}{3pt}
\begin{tabular}{lrrrr}
\toprule
\tableheadrow
\thead{Variant} & \thead{Clean} & \thead{1.5\%} & \thead{2.5\%} & \thead{5\%}\\
\midrule
Standard & 122.77 & 141.10 & 155.51 & 191.67\\
\qknorm & 111.77 & 129.56 & 143.63 & 177.74\\
QUEST & 116.07 & 134.38 & 148.94 & 184.26\\
\methodrow
\method & \textbf{95.60} & \textbf{112.12} & \textbf{124.12} & \textbf{153.27}\\
\bottomrule
\end{tabular}
\caption{Perplexity on clean and word-replaced WikiText-103 at 190M.}
\label{tab:text-perturbation}
\end{minipage}\hfill
\begin{minipage}[t]{0.45\linewidth}
\vspace{0pt}
\method{} achieves the lowest perplexity on clean text and at every
tested replacement rate (Table~\ref{tab:text-perturbation}).
At 5\% replacement, its perplexity is 13.8\% below \qknorm{} and
16.8\% below QUEST.
Its language-modeling advantage therefore persists as local
word information is disrupted.
\end{minipage}
\end{table}

\section{Related Work}

\paragraph{Attention Scaling and Normalization.}
Scaled dot-product attention divides logits by $\sqrt{d_h}$, which
controls their scale at initialization but does not bound later growth
\citep{vaswani2017attention}.
Transformer stability has also been studied through normalization
placement and residual scaling \citep{xiong2020layernorm}.
ScaleNorm, NormFormer, and DeepNorm provide further normalization- and
residual-based designs
\citep{nguyen2019scalenorm,shleifer2021normformer,wang2022deepnet}.
Other studies trace training failures to loss spikes, growing attention
logits, or entropy collapse
\citep{takase2023spike,dehghani2023scaling,zhai2023entropy,
wortsman2023proxies}.
These methods mainly stabilize the wider Transformer block.
A more direct line controls the attention scores themselves.
Examples include Lipschitz-normalized attention
\citep{kim2021lipschitz,dasoulas2021lipschitz}, scaled cosine attention
\citep{henry2020qknorm,liu2022swinv2}, logit soft capping
\citep{gemmateam2024gemma2,rybakov2024stability}, and methods that limit
changes in attention logits during optimization
\citep{anson2025controlling}.
QK-Norm is now used in several large-model architectures
\citep{chameleonteam2024chameleon,teamolmo2025olmo2,
teamolmo2025olmo3,yang2025qwen3,gemmateam2025gemma3}, and has also been
adapted to latent-attention designs \citep{han2026qknormmla}.

These controls improve stability, but they also change which input
properties can affect attention.
QKNorm replaces raw dot products with scaled cosine similarity
\citep{henry2020qknorm}, so radial changes in a projected query or key are
removed.
Recent work has started to preserve selected magnitude effects.
QUEST keeps a query-dependent temperature while normalizing keys
\citep{govindarajan2026quest}, while Norm$\times$Direction restores query
magnitude in linear attention \citep{meng2025normdirection}.
QK-MAP addresses the same broad trade-off from a different direction:
it keeps both query- and key-side input changes available through a
shared bounded representation, then learns separate query and key
projections.

\paragraph{Periodic representations in neural networks.}
Periodic features have been used for several different goals.
Random Fourier features approximate shift-invariant kernels
\citep{rahimi2007random}, and Performer uses positive random features to
approximate softmax attention \citep{choromanski2021performer}.
Fourier input mappings and periodic activations improve the
representation of high-frequency signals
\citep{tancik2020fourier,sitzmann2020siren}.
Learned Fourier features encode spatial coordinates
\citep{li2021learnablefourier}, while RoPE uses paired rotations to encode
relative position in attention \citep{su2024roformer}.
Time2Vec similarly maps a scalar into learned linear and periodic
coordinates \citep{kazemi2019time2vec}.
FAN further combines learned sine--cosine features with non-periodic
features as a general network component \citep{dong2025fan}.
QK-MAP uses a related paired construction, but its purpose is different:
it does not approximate an attention kernel or encode token position.
Instead, it learns a bounded multi-frequency representation of the input
to the query and key projections.

\section{Conclusion}

QK-Norm controls attention-logit growth, but also removes much of the
token-specific magnitude information carried by queries and keys.
We introduced \method{}, which uses learned multi-frequency periodic
features to represent magnitude without relying on feature-norm growth.
Experiments from 190M to 7B show consistent improvements in language
modeling.
Attention analyses further show lower first-key attention and stronger
attention contrast for relevant evidence.
Overall, our results show that query-key scale can be controlled without
discarding magnitude information, making \method{} a practical
alternative to QK normalization.

\subsection*{AI use statement}
In preparing this manuscript, we use LLMs to aid and polish the writing.
Specifically, LLMs improve clarity, grammar, and phrasing, ensuring the
text is concise and readable. The use of LLMs does not influence the
technical contributions or the interpretation of experimental findings.
All content polished by LLMs is carefully checked by the authors.

\subsection*{Reproducibility statement}
We provide anonymous source code for the QK-MAP module, OLMo integration
helpers, training-configuration scaffolding, analysis probes, and
evaluation scripts at \url{https://github.com/yongding-tao/QK-MAP}.
Matched training settings, residual options, and compute notes are given
in Appendix~\ref{app:configuration}. Downstream metrics, validation
corpora, and text-perturbation protocols are specified in
Appendices~\ref{app:evaluation} and~\ref{app:additional-probes}.
Proofs of radial sensitivity and bounded logits appear in
Appendix~\ref{app:proofs}, and architecture and residual ablations in
Appendix~\ref{app:ablations}. Within each scale, all variants share the
same backbone, data, token budget, batch size, and seeds, as described
in Section~\ref{sec:experiments}.

\bibliography{references}
\clearpage

\appendix
\setcounter{proposition}{0}
\renewcommand{\theproposition}{A.\arabic{proposition}}
\providecommand{\theHproposition}{}
\renewcommand{\theHproposition}{appendix.\arabic{proposition}}
\startcontents[appendix]
\printcontents[appendix]{l}{1}{\section*{Appendix Contents}}
\clearpage
\section{Theoretical Details}
\label{app:proofs}

QK-MAP has two design goals: preserving information about token magnitude
and keeping attention logits bounded as that magnitude grows. We analyze
both properties below.

\subsection{Setup and Assumptions}

We use column vectors. For a query token at position $i$, write
$x_i=r_i u_i$, where $r_i>0$ is its magnitude and $u_i\neq0$ is a fixed
direction. A key token is written in the same way as $y_j=t_j v_j$.
The query and key projections are bias-free, as in our implementation.
The RoPE matrices $R_i$ and $R_j$ are orthogonal, so they preserve Euclidean
norms. The scaled attention logit is
\begin{equation}
s_{ij}=\frac{q_i^\top k_j}{\sqrt{d_h}}.
\label{eq:supp-attention-score}
\end{equation}
For QK-RMSNorm, let
\begin{equation}
\mathcal{N}_{\epsilon,G}(z)
=
\frac{Gz}{\sqrt{\|z\|_2^2/d_h+\epsilon}},
\label{eq:supp-rmsnorm}
\end{equation}
where $G$ is the diagonal learned gain and $\epsilon\geq0$. The gain is
shared across tokens and has finite operator norm.

\subsection{Radial Information Lost by QK-RMSNorm}

\begin{proposition}[Radial invariance and score sensitivity]
\label{prop:supp-qknorm-scale}
Let $a=W_Q u_i\neq0$, $c_a=\|a\|_2^2/d_h$, and
\[
q_i(r_i)=R_i\mathcal{N}_{\epsilon,G_Q}(r_i a).
\]
For idealized normalization ($\epsilon=0$), both $q_i(r_i)$ and
$s_{ij}$ are invariant to every positive rescaling of $r_i$. For
$\epsilon>0$,
\begin{align}
\left\|\frac{\partial q_i}{\partial r_i}\right\|_2
&\leq
\frac{\epsilon\|G_Q\|_{\mathrm{op}}\|a\|_2}
     {(r_i^2c_a+\epsilon)^{3/2}},
\label{eq:supp-query-bound}\\
\left|\frac{\partial s_{ij}}{\partial r_i}\right|
&\leq
\frac{\epsilon\|G_Q\|_{\mathrm{op}}\|G_K\|_{\mathrm{op}}\|a\|_2}
     {(r_i^2c_a+\epsilon)^{3/2}}.
\label{eq:supp-score-bound}
\end{align}
An analogous bound holds for the key magnitude $t_j$. Once the projected
query norm is outside the small-$\epsilon$ regime, the radial sensitivity of
QK-RMSNorm decays as $O(r_i^{-3})$.
\end{proposition}

\begin{proof}
Substituting $W_Qx_i=r_i a$ into
Equation~\ref{eq:supp-rmsnorm} gives
\begin{equation}
q_i(r_i)
=R_iG_Q\frac{r_i a}{\sqrt{r_i^2c_a+\epsilon}}.
\label{eq:supp-query-radial}
\end{equation}
When $\epsilon=0$ and $r_i>0$, the scale cancels:
\begin{equation}
q_i(r_i)=R_iG_Q\frac{a}{\sqrt{c_a}}.
\end{equation}
The query is therefore identical for all positive $r_i$. Holding the key
fixed in Equation~\ref{eq:supp-attention-score} makes the score invariant
as well.

For $\epsilon>0$, differentiating the scalar factor in
Equation~\ref{eq:supp-query-radial} yields
\begin{align}
\frac{\partial q_i}{\partial r_i}
&=R_iG_Qa\left[
\frac{1}{(r_i^2c_a+\epsilon)^{1/2}}
-\frac{r_i^2c_a}{(r_i^2c_a+\epsilon)^{3/2}}
\right]\\
&=R_iG_Qa\,
\frac{\epsilon}{(r_i^2c_a+\epsilon)^{3/2}}.
\label{eq:supp-query-derivative}
\end{align}
Because $R_i$ is orthogonal,
$\|R_iG_Qa\|_2\leq\|G_Q\|_{\mathrm{op}}\|a\|_2$, which proves
Equation~\ref{eq:supp-query-bound}.

It remains to transfer this result to the attention score. For any $z$,
\begin{align}
\|\mathcal{N}_{\epsilon,G_K}(z)\|_2
&\leq
\|G_K\|_{\mathrm{op}}
\frac{\|z\|_2}{\sqrt{\|z\|_2^2/d_h+\epsilon}}\\
&\leq \sqrt{d_h}\|G_K\|_{\mathrm{op}}.
\end{align}
RoPE preserves this bound, so $\|k_j\|_2\leq
\sqrt{d_h}\|G_K\|_{\mathrm{op}}$. The Cauchy--Schwarz inequality then gives
\begin{align}
\left|\frac{\partial s_{ij}}{\partial r_i}\right|
&=
\frac{1}{\sqrt{d_h}}
\left|\left(\frac{\partial q_i}{\partial r_i}\right)^\top k_j\right|\\
&\leq
\frac{1}{\sqrt{d_h}}
\left\|\frac{\partial q_i}{\partial r_i}\right\|_2\|k_j\|_2,
\end{align}
which, together with Equation~\ref{eq:supp-query-bound}, proves
Equation~\ref{eq:supp-score-bound}. Interchanging query and key proves the
corresponding statement for $t_j$.
\end{proof}

The proposition concerns the attention score itself, rather than only an
intermediate normalized vector. A learned gain changes the constant in the
bound, but cannot remove the exact invariance at $\epsilon=0$ or the cubic
decay for finite $\epsilon$.

\subsection{Bounded and Magnitude-Sensitive QK-MAP}

Let $W_p=[w_1,\ldots,w_m]\in\mathbb{R}^{D\times m}$ and
$b\in\mathbb{R}^m$. QK-MAP uses
\begin{equation}
\theta(x)=W_p^\top x+b,\qquad
\Phi(x)=
\begin{bmatrix}
\cos\theta(x)\\
\sin\theta(x)
\end{bmatrix}\in\mathbb{R}^{2m}.
\label{eq:supp-feature-map}
\end{equation}

\begin{proposition}[Bounded features and directional sensitivity]
\label{prop:supp-qkmap-scale}
For every $x\in\mathbb{R}^{D}$,
\begin{equation}
\|\Phi(x)\|_2^2=m.
\label{eq:supp-map-norm}
\end{equation}
Its Jacobian satisfies
\begin{equation}
J_\Phi(x)^\top J_\Phi(x)=W_pW_p^\top.
\label{eq:supp-jacobian-gram}
\end{equation}
Consequently, for any perturbation $\delta$,
\begin{equation}
\|J_\Phi(x)\delta\|_2^2=\|W_p^\top\delta\|_2^2.
\label{eq:supp-direction-response}
\end{equation}
In particular, along $x=ru$,
\begin{equation}
\left\|\frac{\partial\Phi(ru)}{\partial r}\right\|_2
=\|W_p^\top u\|_2,
\label{eq:supp-qkmap-scale-response}
\end{equation}
which is positive whenever $u\notin\ker(W_p^\top)$ and does not decay as
$r$ grows.
\end{proposition}

\begin{proof}
Writing $\theta_j=w_j^\top x+b_j$, each paired coordinate obeys
$\cos^2\theta_j+\sin^2\theta_j=1$. Summing the $m$ pairs proves
Equation~\ref{eq:supp-map-norm}. Elementwise differentiation gives
\begin{equation}
J_\Phi(x)=
\begin{bmatrix}
-\operatorname{Diag}(\sin\theta)W_p^\top\\
\phantom{-}\operatorname{Diag}(\cos\theta)W_p^\top
\end{bmatrix}.
\end{equation}
Therefore,
\begin{align}
J_\Phi(x)^\top J_\Phi(x)
&=W_p\operatorname{Diag}(\sin^2\theta)W_p^\top\\
&\quad+W_p\operatorname{Diag}(\cos^2\theta)W_p^\top\\
&=W_pW_p^\top,
\end{align}
which proves Equation~\ref{eq:supp-jacobian-gram}. Multiplying on the left
and right by $\delta^\top$ and $\delta$ proves
Equation~\ref{eq:supp-direction-response}. Setting $\delta=u$ gives the
radial result.
\end{proof}

\begin{proposition}[Magnitude-independent logit bound]
\label{prop:supp-map-logit-bound}
Let
\[
q_i=R_iA_Q\Phi(x_i),\qquad k_j=R_jA_K\Phi(y_j),
\]
where $A_Q$ and $A_K$ are learned linear projections. Then
\begin{equation}
|s_{ij}|
\leq
\frac{m}{\sqrt{d_h}}
\|A_Q\|_{\mathrm{op}}\|A_K\|_{\mathrm{op}},
\label{eq:supp-map-score-bound}
\end{equation}
independently of $\|x_i\|_2$ and $\|y_j\|_2$. The mapped query remains
radially sensitive whenever
$A_QJ_\Phi(x_i)u_i\neq0$, with the analogous condition for the key.
\end{proposition}

\begin{proof}
Equation~\ref{eq:supp-map-norm} and orthogonality of RoPE give
\[
\|q_i\|_2\leq\sqrt{m}\|A_Q\|_{\mathrm{op}},
\qquad
\|k_j\|_2\leq\sqrt{m}\|A_K\|_{\mathrm{op}}.
\]
Applying Cauchy--Schwarz to Equation~\ref{eq:supp-attention-score} proves
Equation~\ref{eq:supp-map-score-bound}. The chain rule gives
$\partial q_i/\partial r_i=R_iA_QJ_\Phi(x_i)u_i$, which is nonzero under
the stated condition.
\end{proof}

The default implementation adds a learned residual,
\begin{equation}
\Psi_\alpha(x)=\Phi(x)+\operatorname{Diag}(\alpha)x,
\label{eq:supp-residual-map}
\end{equation}
with $2m=D$ and $\alpha$ initialized at zero. The attention-block input is
produced by a preceding RMSNorm. If $\|x\|_2\leq B$, then
\begin{equation}
\|\Psi_\alpha(x)\|_2
\leq\sqrt{m}+\|\alpha\|_\infty B.
\label{eq:supp-residual-bound}
\end{equation}
Thus, the residual variant starts exactly as the bounded periodic map and
remains bounded on the normalized input domain. This connects the theoretical
map to the residual design used in the scale experiments.

\subsection{When Different Magnitudes Produce the Same Features}

For a fixed direction $u$, define radial frequencies
$\omega_j=w_j^\top u$. Equality of the paired sine and cosine features is
equivalent to
\begin{equation}
(r_1-r_2)\omega_j\in2\pi\mathbb{Z}
\qquad\text{for every }j.
\label{eq:supp-collision}
\end{equation}

\begin{proposition}[A sufficient condition for global injectivity]
\label{prop:supp-injectivity}
Suppose two nonzero radial frequencies satisfy
$\omega_p/\omega_q\notin\mathbb{Q}$. Then the map
$r\mapsto\Phi(ru)$ is injective on $\mathbb{R}$.
\end{proposition}

\begin{proof}
Assume that $r_1\neq r_2$ produces a collision and let
$\Delta=r_1-r_2\neq0$. Equation~\ref{eq:supp-collision} implies
$\Delta\omega_p=2\pi n_p$ and $\Delta\omega_q=2\pi n_q$ for integers
$n_p,n_q$. Since $\Delta$, $\omega_p$, and $\omega_q$ are nonzero, both
integers are nonzero. Dividing the two equalities gives
$\omega_p/\omega_q=n_p/n_q\in\mathbb{Q}$, contradicting the assumption.
\end{proof}

If $W_p$ is drawn from a jointly absolutely continuous
distribution, the rational-ratio failure set has measure zero for any fixed
nonzero $u$ and $m\geq2$. Hence, the proposition holds almost surely at
initialization. Learning may change the frequencies, so this is a sufficient
condition rather than an unconditional guarantee.

\begin{proposition}[Separation on a bounded magnitude range]
\label{prop:supp-finite-separation}
Assume the condition of Proposition~\ref{prop:supp-injectivity}. For any
compact interval $I\subset\mathbb{R}$ and any $\eta>0$, there exists
$\gamma(I,\eta)>0$ such that
\begin{equation}
\|\Phi(r_1u)-\Phi(r_2u)\|_2\geq\gamma(I,\eta)
\end{equation}
for all $r_1,r_2\in I$ with $|r_1-r_2|\geq\eta$.
\end{proposition}

\begin{proof}
The set
$K=\{(r_1,r_2)\in I^2:|r_1-r_2|\geq\eta\}$ is compact.
If $K$ is empty, the claim holds for any $\gamma>0$.
Otherwise, the function
$f(r_1,r_2)=\|\Phi(r_1u)-\Phi(r_2u)\|_2$ is continuous. By injectivity,
$f$ is strictly positive on $K$. The extreme-value theorem therefore gives
a positive attained minimum $\gamma(I,\eta)=\min_K f$.
\end{proof}

This final result states the practical role of multiple frequencies: on any
finite magnitude range, inputs separated by a fixed amount also have a
strictly positive feature separation.

\section{Experimental Setup}

We compare Standard attention, QK-Norm, and QK-MAP at 190M, 370M, 760M,
1B, and 7B parameters. The five scales are trained for 2B, 4B, 8B, 10B,
and 20B tokens, respectively. All runs use sequence length 2048 and a
global batch size of 2,097,152 tokens.

Within each scale, the three variants share the layer count, hidden size,
attention-head configuration, tokenizer, pretraining mixture, optimizer,
token budget, global batch size, initialization seed, and data order. The
pretraining corpus is the same Dolma-based OLMo mixture for every variant.

QK-MAP adds a periodic projection. We reduce its feed-forward width so that
the total parameter count remains closely matched to Standard attention.
The difference is at most 0.033\% at any scale and 0.017\% at 7B.

The 190M architecture metadata were reconstructed from the archived training
code and records.

\subsection{Downstream Evaluation}

Avg-8 is the unweighted mean accuracy over ARC-Easy, PIQA, SciQ,
ARC-Challenge, COPA, HellaSwag, OpenBookQA, and WinoGrande. All eight tasks
are evaluated zero-shot.

The common evaluation checkpoints for 190M, 370M, 760M, 1B, and 7B are
steps 954, 1908, 3815, 4769, and 9465, respectively. All variants within a
scale use the same checkpoint and evaluation settings.

\subsection{Nearby-Checkpoint Evaluation}
\label{app:checkpoint-robustness}

We also evaluate 16 common 7B checkpoints from steps 9455 to 9537,
shown in Figure~\ref{fig:7b-dynamics}(c).
QK-MAP ranks first at every checkpoint. Its Avg-8 margin over Standard
averages 0.45 percentage points and ranges from 0.06 to 0.90 points.
The consistent ordering across nearby endpoints shows that the result is not
caused by selecting one favorable checkpoint.

\section{Detailed Configurations and Reproducibility}
\label{app:configuration}

This section gives the scale-dependent architectures, parameter matching,
shared optimization settings, and exact evaluation protocol needed to
reproduce the experiments.

\subsection{Scale-Specific Configurations and Parameter Matching}

Values for 370M--7B come from exact resolved run configurations.
The 190M metadata were reconstructed from archived training code.

\begin{table}[!tbp]
\centering
\small
\renewcommand{\arraystretch}{1.08}
\setlength{\tabcolsep}{4.5pt}
\begin{fittable}
\begin{tabular}{lrrrrrrrrr}
\toprule
\tableheadrow
\thead{Scale} & \thead{$L$} & \thead{$D$} & \thead{$H$} &
\thead{$d_h$} & \thead{FF$_{\mathrm{base}}$} &
\thead{FF$_{\mathrm{MAP}}$} & \thead{Tokens} & \thead{Steps} &
\thead{Peak LR}\\
\midrule
190M & 12 & 768  & 12 & 64  & 3072  & 2944  & 2B  & 954  & $6{\times}10^{-4}$\\
370M & 16 & 1024 & 16 & 64  & 4096  & 3928  & 4B  & 1908 & $6{\times}10^{-4}$\\
760M & 16 & 1536 & 16 & 96  & 6144  & 5888  & 8B  & 3815 & $6{\times}10^{-4}$\\
1B   & 18 & 2048 & 16 & 128 & 8192  & 7852  & 10B & 4769 & $4{\times}10^{-4}$\\
7B   & 32 & 4096 & 32 & 128 & 11008 & 10328 & 20B & 9537 & $3{\times}10^{-4}$\\
\bottomrule
\end{tabular}
\end{fittable}
\caption{Backbone and schedule by scale.
$L$ is the number of layers, $D$ the hidden size, $H$ the number of heads,
and $d_h$ the head width.
FF$_{\mathrm{base}}$ applies to Standard and \qknorm{}; FF$_{\mathrm{MAP}}$
is the reduced \method{} feed-forward width used for parameter matching.
All runs use sequence length 2048 and global batch size 2,097,152 tokens.}
\label{tab:scale-config}
\end{table}

\begin{table}[!tbp]
\centering
\small
\renewcommand{\arraystretch}{1.08}
\setlength{\tabcolsep}{8pt}
\begin{fittable}
\begin{tabular}{lrrr@{\hspace{10pt}}r}
\toprule
\tableheadrow
\thead{Scale} & \thead{Standard} & \thead{\qknorm} &
\thead{\method} & \thead{\shortstack{$\Delta$ params\\vs.\ Standard}}\\
\midrule
190M & 267,406,080 & 267,424,512 & \methodcell{267,419,904} & +0.0052\%\\
370M & 473,990,144 & 474,022,912 & \methodcell{474,145,792} & +0.0328\%\\
760M & 912,311,808 & 912,360,960 & \methodcell{912,348,672} & +0.0040\%\\
1B   & 1,619,077,120 & 1,619,150,848 & \methodcell{1,619,279,872} & +0.0125\%\\
7B   & 7,298,355,200 & 7,298,617,344 & \methodcell{7,299,600,384} & +0.0171\%\\
\bottomrule
\end{tabular}
\end{fittable}
\caption{Analytical parameter counts from the resolved 370M--7B
configurations and reconstructed 190M metadata.
The Q/K RMSNorm and periodic-map parameters produce small differences.
\method{} remains within 0.033\% of Standard at every reported scale;
the final column reports its relative parameter increase.}
\label{tab:parameter-counts}
\end{table}

Tables~\ref{tab:scale-config} and~\ref{tab:parameter-counts} isolate the
attention formulation while keeping the backbone, training budget, and total
parameter count closely aligned.
The largest parameter difference from Standard is 0.033\%, and the difference
at the largest scale is 0.017\%.
Thus, a material increase in model size is unlikely to explain the reported
optimization and downstream differences.

\subsection{Shared Optimization and Implementation}

\begin{table}[!tbp]
\centering
\small
\renewcommand{\arraystretch}{1.08}
\setlength{\tabcolsep}{5pt}
\begin{fittable}
\begin{tabular}{llll}
\toprule
\tableheadrow
\thead{Field} & \thead{Value} & \thead{Field} & \thead{Value}\\
\midrule
Optimizer & AdamW &
RoPE theta & 500,000\\
Betas / epsilon & $(0.9,0.95)$ / $10^{-8}$ &
Tokenizer & \texttt{allenai/dolma2-tokenizer}\\
Weight decay / grad clip & 0.1 / 1.0 &
Vocabulary / padded size & 100,278 / 100,352\\
Warmup / decay & 5\% / cosine &
Data mixture & \texttt{OLMo-mix-0625-20Bsample}\\
Final LR ratio & 0.1 &
Experiment / model / data seeds & 12536 / 0 / 34521\\
Z-loss multiplier & $10^{-5}$ &
\qknorm{} epsilon & $10^{-6}$\\
Parameter / reduction dtype & BF16 / FP32 &
\methodcell{\method{} periodic ratio} & \methodcell{0.5}\\
\methodcell{\method{} residual init} & \methodcell{per-dimension, zero} &
\methodcell{\method{} map scope} & \methodcell{shared by Q and K}\\
\bottomrule
\end{tabular}
\end{fittable}
\caption{Settings shared across the reported scales and variants.}
\label{tab:shared-config}
\end{table}

Together, the shared settings in Table~\ref{tab:shared-config} make each
within-scale comparison matched in optimizer, precision, tokenizer, data
order, and initialization seed.
This pairing isolates the attention formulation while keeping the main
sources of training variation fixed.

\subsection{Released Implementation and Compute}
\label{app:reproducibility}

The code package contains the QK-MAP module, OLMo integration helpers,
training configuration code, magnitude-response and norm-shuffling probes,
and OLMES aggregation and bootstrap scripts. The implementation documents tensor
shapes, initialization, residual options, and the order of Q/K projection and
RoPE. Pretraining was run with eight NVIDIA A100 40GB GPUs using distributed
data parallelism. The same software stack and precision settings were used
for all variants within each scale.

\section{Evaluation Protocol}
\label{app:evaluation}

All tasks are evaluated zero-shot. The random-subsampling seed is 1234.
The maximum evaluation size is 1000 examples, except that all 500 OpenBookQA
validation examples are used.

\begin{table}[!tbp]
\centering
\small
\renewcommand{\arraystretch}{1.08}
\setlength{\tabcolsep}{5pt}
\begin{fittable}
\begin{tabular}{lcr@{\hspace{34pt}}lcr}
\toprule
\tableheadrow
\thead{Task} & \thead{Metric} & \thead{$N$} &
\thead{Task} & \thead{Metric} & \thead{$N$}\\
\midrule
ARC-Easy & Acc.-unc. & 570 &
ARC-Challenge & Acc.-unc. & 299\\
PIQA & Acc./token & 1000 &
COPA & Accuracy & 100\\
SciQ & Accuracy & 1000 &
OpenBookQA & Acc.-unc. & 500\\
HellaSwag & Acc./token & 1000 &
WinoGrande & Acc./token & 1000\\
\bottomrule
\end{tabular}
\end{fittable}
\caption{Downstream evaluation protocol. All tasks use the validation split
and zero-shot prompting. Acc.-unc.\ denotes unconditioned accuracy;
Acc./token denotes length-normalized accuracy.}
\label{tab:evaluation-protocol}
\end{table}

Table~\ref{tab:evaluation-protocol} fixes the split, prompting, subsampling,
and metric for the eight downstream tasks before comparing model variants.
Avg-8 averages the eight task accuracies without weighting by dataset size.
This gives equal weight to each capability measured by the benchmark suite.

For validation CE, Dolma-6 is the mean over Books, Common Crawl, PeS2o, Reddit,
Stack, and Wiki.
The main table's Val.\ column averages C4, Pile, M2D2-S2ORC, and
Dolma-6 with equal weight.
All methods use the same validation shards and evaluator within each scale.

\section{Architecture and Residual Ablations}
\label{app:ablations}

\mapdim{} uses a same-dimensional periodic map without a residual.
\mapparam{} adjusts the map to match parameters without the residual.
\mapres{} is the parameter-matched map with a learned per-dimension residual
used as the default QK-MAP design in the cross-scale experiments.
At 370M, we also compare against \qkclip{}, the optimizer-side
logit-control technique introduced in Kimi K2 \citep{team2025kimi}.
All averages below are computed from the task-level evaluation outputs.

\FloatBarrier
\begin{appendixwidetable}
\centering
\small
\renewcommand{\arraystretch}{1.10}
\setlength{\tabcolsep}{3.4pt}
\begin{fittable}
\begin{tabular}{llccccccccc}
\toprule
\tableheadrow
\thead{Scale} & \thead{Variant} & \thead{ARC-E} & \thead{PIQA} &
\thead{SciQ} & \thead{ARC-C} & \thead{COPA} &
\thead{HS} & \thead{OBQA} & \thead{WG} & \thead{Avg-8}\\
\midrule
370M & Standard & 39.12 & 58.90 & 67.00 & \textbf{31.10} &
60.00 & 25.90 & 28.80 & 52.00 & 45.35\\
& \qknorm & 40.70 & 59.40 & 70.00 & 23.75 &
\textbf{64.00} & 27.00 & 33.60 & 48.80 & 45.91\\
& \qkclip & 38.25 & 59.00 & 68.00 & 29.10 &
\textbf{64.00} & 27.30 & 30.40 & 47.60 & 45.46\\
\subtlerow
& \mapdim & 40.70 & \textbf{60.30} & 71.20 & 28.09 &
61.00 & \textbf{27.50} & 31.40 & \textbf{52.70} &
\textbf{46.61}\\
\subtlerow
& \mapparam & \textbf{41.05} & 58.80 & \textbf{72.10} &
25.75 & 61.00 & 27.20 & 33.80 & 48.90 & 46.08\\
\methodrow
& \mapres & 40.70 & 58.00 & 69.70 & 25.42 & 61.00 &
27.30 & \textbf{35.60} & 49.80 & 45.94\\
\midrule
760M & Standard & 47.72 & 62.00 & 79.80 & \textbf{31.44} &
64.00 & 32.30 & 34.40 & 50.30 & 50.24\\
& \qknorm & 47.19 & 61.70 & 77.50 & 29.77 &
\textbf{65.00} & 32.30 & 33.60 & 49.80 & 49.61\\
\subtlerow
& \mapdim & 45.09 & \textbf{63.30} & \textbf{81.00} &
30.10 & 64.00 & 32.90 & \textbf{38.60} & 51.30 & \textbf{50.79}\\
\subtlerow
& \mapparam & \textbf{47.89} & 63.00 & 80.20 & 31.10 &
63.00 & 32.20 & 36.40 & \textbf{51.60} & 50.67\\
\methodrow
& \mapres & 45.09 & 62.50 & 79.90 & 28.43 & 64.00 &
\textbf{33.20} & 34.80 & 49.90 & 49.73\\
\midrule
1B & Standard & 48.42 & \textbf{63.90} & \textbf{82.00} &
27.76 & 66.00 & 35.00 & 36.00 & \textbf{51.70} & 51.35\\
& \qknorm & \textbf{51.05} & 63.40 & 79.40 & 31.77 &
66.00 & 35.70 & \textbf{37.40} & 51.50 & 52.03\\
\subtlerow
& \mapdim & 47.72 & \textbf{63.90} & 80.10 &
32.78 & 66.00 & 33.30 & 34.60 & 50.10 & 51.06\\
\subtlerow
& \mapparam & 50.18 & 63.50 & 80.60 & 31.44 &
\textbf{67.00} & \textbf{36.40} & 35.60 & 50.00 & 51.84\\
\methodrow
& \mapres & 49.30 & \textbf{63.90} & \textbf{82.00} &
\textbf{34.78} & \textbf{67.00} & 34.20 & 34.40 & 51.30 &
\textbf{52.11}\\
\bottomrule
\end{tabular}
\end{fittable}
\caption{Downstream architecture ablations at the matched checkpoint for each
scale. All entries are accuracies (\%), and Avg-8 is computed from the eight
displayed tasks.
\mapdim{} and \mapparam{} omit the residual, whereas \mapres{} is the
parameter-matched residual implementation used as the default \method{} design.
\qkclip{} was evaluated only at 370M.
Best values within each scale are bold.}
\label{tab:ablation-summary}
\end{appendixwidetable}

Table~\ref{tab:ablation-summary} compares these variants at 370M, 760M, and 1B.
At each scale, at least one QK-MAP variant outperforms both Standard
attention and QK-Norm. The direct map \mapdim{} leads at 370M and 760M, while the
parameter-matched residual map \mapres{} leads among periodic variants at
1B. This heterogeneous ablation shows that the gain is not tied to a single
integration rule. It also supports the residual design at larger scale,
where the periodic branch can refine the original representation.

\FloatBarrier
\section{Attention Diagnostics}
\label{app:diagnostics}

\subsection{Synthetic Evidence Contrast}
\label{app:evidence-contrast}

The main comparison in Table~\ref{tab:niah-average-layer} averages
the last-query evidence attention SNR over all layers.
The query is the final question token; inserted evidence and the
surrounding distractor tokens define the signal and noise regions.
The same masks and layer-pooling rule are used for all variants.

The single-needle set contains 112,200 examples: 600 examples per
setting, eleven depth layouts, and seventeen sequence lengths from
32 to 600 tokens.
The two- and three-needle sets contain 201,600 and 162,000 examples,
respectively, and the mixed-task set contains 112,200 examples.
All sets use seed 20251231 and the 7B step-9465 checkpoints.

\begin{figure}[!tbp]
\centering
\includegraphics[width=0.96\linewidth]{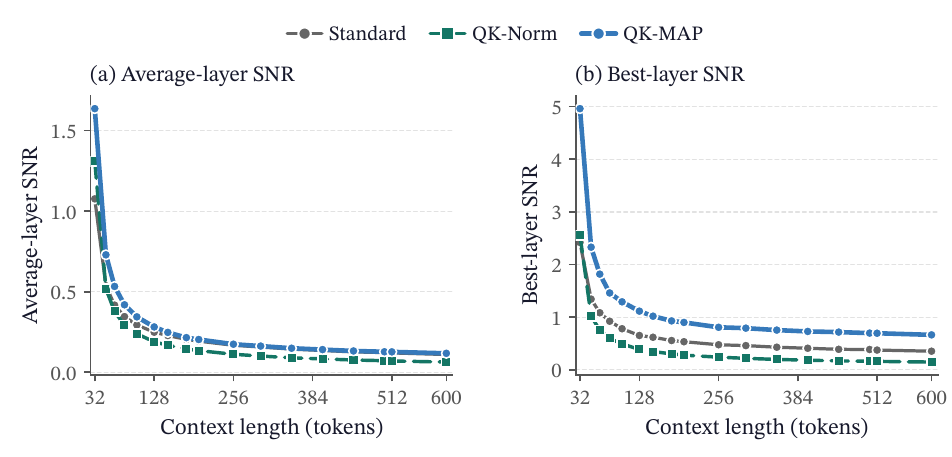}
\caption{Single-needle attention SNR under two layer-pooling rules.
(a) Average across layers. (b) Best layer for each example.
Each point aggregates 6,600 examples.}
\label{fig:niah}
\end{figure}

Figure~\ref{fig:niah} compares two layer-pooling rules.
Average-layer SNR uses every layer for each example.
Best-layer SNR takes the largest layer score within each example
before averaging over examples.
The main text uses the average-layer result, and the best-layer view
shows that the ordering also holds for the strongest layer response.
\method{} leads both baselines at every evaluated length under both
pooling rules.

\subsection{Cross-Domain First-Token Allocation}
\label{app:first-token-allocation}

For the 7B checkpoint, we measure the percentage of attention assigned to
the first key, averaged over layers, heads, and query positions:
\begin{equation}
\mathrm{F\mbox{-}Attn}
=
\frac{100}{LHT}
\sum_{\ell=1}^{L}\sum_{h=1}^{H}\sum_{i=1}^{T}A_h^\ell[i,1].
\label{eq:fattn-appendix}
\end{equation}
The statistic is computed on consecutive full-length validation chunks;
incomplete final chunks are discarded. Uniform causal attention has
reference value $100\mathcal{H}_T/T$, where $\mathcal{H}_T$ is the $T$th
harmonic number.

\begin{appendixwidetable}
\centering
\begin{minipage}{\textwidth}
\centering
\footnotesize
\renewcommand{\arraystretch}{1.07}
\setlength{\tabcolsep}{2.2pt}
\begin{fittable}
\begin{tabular}{clcccccccccccc}
\toprule
\tableheadrow
\thead{$T$} & \thead{Variant} & \thead{C4} & \thead{Books} &
\thead{CC} & \thead{PeS2o} & \thead{Reddit} & \thead{Stack} &
\thead{Wiki} & \thead{ICE} & \thead{M2D2} & \thead{Pile} &
\thead{W103} & \thead{Avg.\ $\downarrow$}\\
\midrule
\multirow{3}{*}{32}
& Standard & 21.91 & 21.36 & 21.71 & 23.79 & 21.89 & 22.39 &
22.71 & 20.86 & 22.65 & 22.97 & 25.54 & 22.53\\
& \qknorm & 19.84 & 20.05 & 19.68 & 21.19 & 20.58 & 22.59 &
20.25 & 20.58 & 21.07 & 21.51 & 24.83 & 21.11\\
\methodrow
& \method & 19.80 & 19.06 & 19.57 & 21.02 & 20.35 & 20.06 &
19.97 & 19.61 & 20.95 & 20.56 & 24.43 & \textbf{20.49}\\
\midrule
\multirow{3}{*}{64}
& Standard & 14.09 & 14.49 & 13.88 & 15.73 & 13.87 & 15.27 &
14.67 & 13.77 & 14.34 & 15.17 & 17.93 & 14.84\\
& \qknorm & 13.20 & 14.07 & 12.88 & 14.23 & 13.32 & 16.25 &
13.37 & 14.17 & 13.82 & 14.80 & 17.26 & 14.31\\
\methodrow
& \method & 12.35 & 12.75 & 12.16 & 13.56 & 12.79 & 13.73 &
12.68 & 12.91 & 12.96 & 13.25 & 17.01 & \textbf{13.29}\\
\midrule
\multirow{3}{*}{128}
& Standard & 8.90 & 9.96 & 8.77 & 10.23 & 8.46 & 10.52 &
9.26 & 9.02 & 8.47 & 9.93 & 11.96 & 9.59\\
& \qknorm & 8.83 & 9.82 & 8.64 & 9.80 & 8.62 & 11.90 &
8.93 & 9.83 & 8.77 & 10.25 & 11.63 & 9.73\\
\methodrow
& \method & 7.64 & 8.63 & 7.54 & 8.65 & 7.73 & 9.57 &
7.82 & 8.50 & 7.53 & 8.54 & 11.41 & \textbf{8.50}\\
\midrule
\multirow{3}{*}{256}
& Standard & 5.52 & 7.03 & 5.47 & 6.59 & 4.92 & 7.26 &
5.80 & 5.86 & 4.73 & 6.51 & 7.62 & 6.12\\
& \qknorm & 5.81 & 6.98 & 5.66 & 6.55 & 5.47 & 8.58 &
5.95 & 6.72 & 5.33 & 7.03 & 7.64 & 6.52\\
\methodrow
& \method & 4.66 & 6.12 & 4.52 & 5.41 & 4.57 & 6.68 &
4.79 & 5.55 & 4.21 & 5.52 & 7.42 & \textbf{5.40}\\
\midrule
\multirow{3}{*}{512}
& Standard & 3.40 & 4.94 & 3.26 & 4.13 & 2.74 & 4.97 &
3.66 & 3.77 & 2.55 & 4.21 & 4.68 & 3.85\\
& \qknorm & 3.84 & 4.76 & 3.46 & 4.25 & 3.24 & 5.85 &
3.91 & 4.48 & 3.04 & 4.71 & 4.79 & 4.21\\
\methodrow
& \method & 2.89 & 4.16 & 2.58 & 3.23 & 2.56 & 4.53 &
2.98 & 3.59 & 2.29 & 3.57 & 4.62 & \textbf{3.36}\\
\bottomrule
\end{tabular}
\end{fittable}
\caption{F-Attn (\%) over eleven validation domains for the 7B step-9465
checkpoint. Lower values indicate less average attention assigned to the
first key; averages are unweighted over domains. Books = Dolma Books;
CC = Dolma Common Crawl; PeS2o = Dolma PeS2o; Stack = Dolma Stack;
Wiki = Dolma Wiki; M2D2 = M2D2-S2ORC; and W103 = WikiText-103.}
\label{tab:fattn-complete}
\end{minipage}
\end{appendixwidetable}

\FloatBarrier
\begin{appendixwidefigure}
\centering
\includegraphics[width=0.92\textwidth]{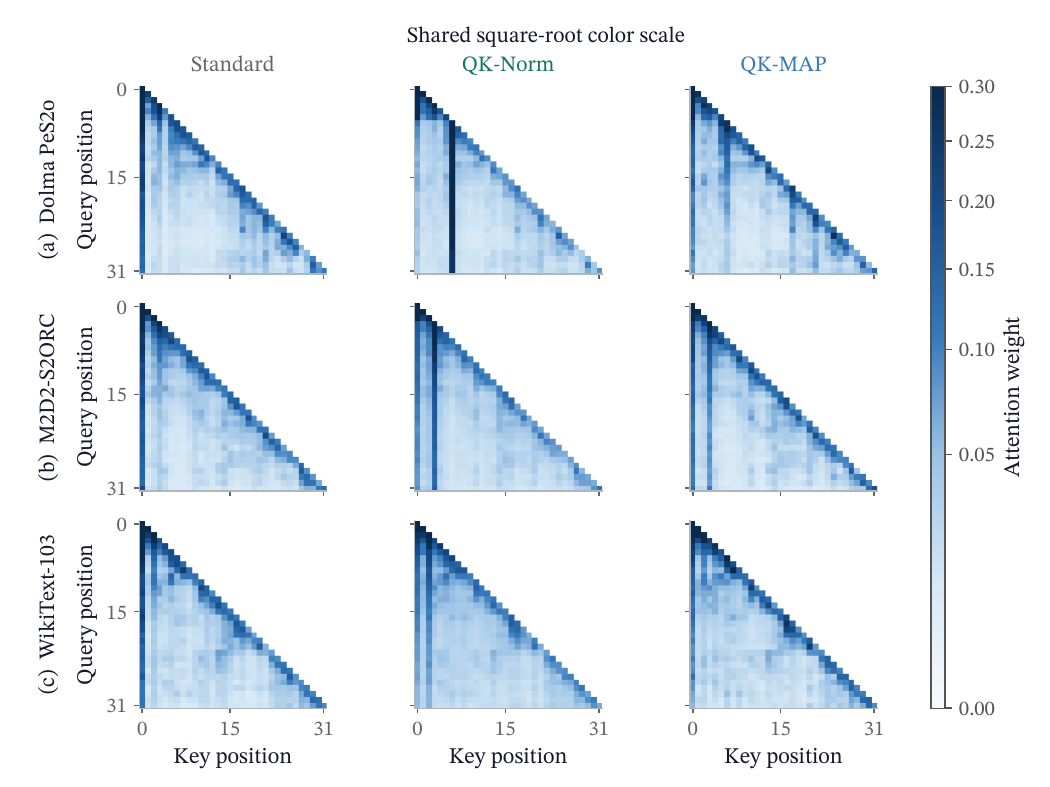}
\caption{Complete causal-attention matrix display for WikiText-103,
M2D2-S2ORC, and Dolma PeS2o at the 7B step-9465 checkpoint.
The nine layer/head-averaged matrices are shown without
domain-level selection and share one square-root color scale.
Across all three domains, QK-MAP shows a weaker first-column band, matching
the lower first-token allocation reported in Table~\ref{tab:fattn-complete}.
Across all 55 domain--length cells, QK-MAP has lower F-Attn than both
baselines; its eleven-domain mean is 9.1--12.7\% lower than Standard across
lengths 32--512.}
\label{fig:attention-heatmaps}
\end{appendixwidefigure}

\FloatBarrier
\section{Magnitude Intervention and Text Perturbation}
\label{app:additional-probes}

\subsection{Norm-Shuffling Protocol}
\label{app:norm-shuffle}

We use three trained 190M checkpoints at approximately 2B tokens:
Standard, QK-RMSNorm, and the same-dimensional direct map \mapdim{}.
These probes use the direct map rather than the parameter-matched
residual variant in Table~\ref{tab:main-results}.
The evaluation sample is the first 128 consecutive 512-token chunks
of the C4 validation token shard, evaluated in FP32 with batch size 16.

\paragraph{Intervention.}
The hook acts after Q/K normalization and RoPE, immediately before
the attention backend, in all twelve layers.
Let $v_{b,t}$ denote the query or key vector for sequence $b$ at
position $t$.
For a permutation $\pi_t$ over the batch, we replace it by
\begin{equation}
\widetilde v_{b,t}
=
\|v_{\pi_t(b),t}\|_2
\frac{v_{b,t}}{\|v_{b,t}\|_2}.
\label{eq:norm-shuffle}
\end{equation}
The implementation clamps denominators below $10^{-12}$.
The permutation is restricted to the same position across sequences,
so the intervention never moves a future-position norm to an
earlier position.
Query-only and key-only conditions change one side; the joint condition
shuffles both.
The main result uses the norm of the full 768-dimensional vector
formed by concatenating twelve heads.
This matches the QK-RMSNorm axis in these checkpoints.

\paragraph{Measurement and controls.}
We report the change in each sequence's CE relative to its unmodified
forward pass, then average across sequences and three permutations.
The seed is 20260725.
We obtain paired 95\% bootstrap intervals by resampling the 128
sequences 10,000 times after averaging repetitions.
The intervals describe variation across this evaluation sample.
The no-hook, identity-hook, and split-and-reconstruct controls agree
to within $9.54\times10^{-7}$ CE per sequence.
This probe evaluates raw contiguous chunks; its clean CE serves only
as the reference for the paired change, rather than replacing the
standard validation evaluator.

\begin{table}[!tbp]
\centering
\small
\renewcommand{\arraystretch}{1.10}
\setlength{\tabcolsep}{5pt}
\begin{fittable}
\begin{tabular}{llrrr}
\toprule
\tableheadrow
\thead{Norm scope} & \thead{Variant} & \thead{Shuffle Q} &
\thead{Shuffle K} & \thead{Shuffle Q and K}\\
\midrule
\multirow{3}{*}{All heads}
& Standard & 0.01651 & 0.18816 & 0.22541\\
& \qknorm & 0.00001 & 0.00012 & 0.00013\\
\methodrow
& \method & 0.00617 & 0.19360 & 0.20106\\
\midrule
\multirow{3}{*}{Per head}
& Standard & 0.17644 & 0.76922 & 1.14959\\
& \qknorm & 0.10227 & 0.62821 & 0.77627\\
\methodrow
& \method & 0.01829 & 0.32990 & 0.37760\\
\bottomrule
\end{tabular}
\end{fittable}
\caption{Mean $\Delta\mathrm{CE}$ for global and headwise norm
shuffling. The global scope matches the QK-RMSNorm axis.}
\label{tab:norm-shuffle-scope}
\end{table}

\begin{table}[!tbp]
\centering
\small
\renewcommand{\arraystretch}{1.10}
\setlength{\tabcolsep}{7pt}
\begin{tabular}{lccc}
\toprule
\tableheadrow
\thead{Variant} & \thead{Shuffle Q} & \thead{Shuffle K} &
\thead{Shuffle Q and K}\\
\midrule
Standard & [0.01503, 0.01802] & [0.18135, 0.19568] & [0.21772, 0.23350]\\
\qknorm & [$-$0.00003, 0.00005] & [$-$0.00009, 0.00033] & [$-$0.00010, 0.00034]\\
\methodrow
\method & [0.00512, 0.00736] & [0.17708, 0.21414] & [0.18283, 0.22208]\\
\bottomrule
\end{tabular}
\caption{Paired 95\% bootstrap intervals for global norm shuffling.}
\label{tab:norm-shuffle-ci}
\end{table}

\paragraph{Normalization axis.}
Table~\ref{tab:norm-shuffle-scope} also reports shuffling each head's
norm separately.
Global normalization removes overall token scale but retains
variation in how magnitude is allocated among heads.
Accordingly, QK-RMSNorm is almost unchanged by global shuffling but
responds to headwise shuffling.
The main comparison targets the global scale that normalization
directly suppresses.
The intervention measures the use of projected query/key magnitudes;
it does not isolate which part of the periodic representation
produced that variation.

\subsection{Word-Replacement Protocol}
\label{app:text-perturbation}

We evaluate Standard, QK-RMSNorm, QUEST, and \mapdim{} at 190M
after approximately 2B pretraining tokens.
We decode the archived WikiText-103 validation token stream,
retain the text through the first 20,000 alphabetic words, and
re-tokenize each perturbed version with the original training tokenizer,
preserving its saved pre-tokenization rules.
For word index $j$, we generate a fixed uniform draw using
seed $20260926+j$ and replace that word by \texttt{AAA} if the
draw is below the chosen rate. Special tokens are left unchanged.
All models receive identical text.
The replacement sets are nested as the rate increases.

The rates 1.5\%, 2.5\%, and 5\% replace 312, 525, and 1,023
words, respectively.
We score every next-token target in causal blocks of at most
512 targets, including the shorter final block.
The clean and perturbed versions contain 28,280, 28,247,
28,220, and 28,156 scored targets.
Perplexity is $\exp(\sum_t \mathrm{CE}_t/N)$ over all scored targets.
This fixed-block implementation differs from QUEST's sliding-window
protocol, so Table~\ref{tab:text-perturbation} is a matched comparison
among our checkpoints, not a reproduction of the original paper's
absolute perplexities.

\method{} has the lowest absolute perplexity at every rate.
At 5\% replacement, the increases relative to each model's own clean
perplexity are 56.1\%, 59.0\%, 58.7\%, and 60.3\% for Standard,
QK-RMSNorm, QUEST, and \method{}, respectively.
The main conclusion therefore concerns the retained absolute
language-modeling advantage under word replacement.

\FloatBarrier

\end{document}